\documentclass[preprint,3p,times,tabularray]{elsarticle}
\usepackage{graphicx}
\usepackage{siunitx}
\usepackage[colorlinks,linkcolor=blue,anchorcolor=blue,citecolor=blue]{hyperref}
\usepackage{array}
\usepackage{amsmath}
\usepackage{tabularx}
\usepackage{float}
\usepackage{subfig}
\usepackage{booktabs}
\usepackage{float}
\usepackage{makecell}
\usepackage{tabularx}
\usepackage{amsmath} 
\usepackage{amssymb}
\usepackage{tikz}
\usepackage{cleveref}
\usepackage{float} 
\usepackage{graphicx}
\usepackage{subfig}
\usepackage{multirow}
\usepackage{caption}
\usepackage{algorithmic}
\usepackage[linesnumbered,ruled]{algorithm2e}

\usepackage{amsmath,amsfonts,amssymb,amsopn,amscd,bbm}
\usepackage{color} 
\usepackage{graphicx} 
\biboptions{numbers,sort&compress}

\journal{Information Sciences}

\begin{document}
\begin{frontmatter}

\title{DPEC: Dual-Path Error Compensation for Low-Light Image Enhancement}

\author[inst1]{Shuang Wang\fnref{equal}} \ead{wangshuang999@scu.edu.cn}
\author[inst2]{Qianwen Lu\fnref{equal}} \ead{396416324@qq.com}
\author[inst1]{Boxing Peng}                         \ead{pbx1035332079@gmail.com}
\author[inst1]{Yihe Nie} \ead{742004668@qq.com}
\author[inst1]{Qingchuan Tao\corref{cor1}} \ead{taoqingchuan@scu.edu.cn}

\fntext[equal]{These authors contributed equally to this work.}
\cortext[cor1]{Corresponding author}
\affiliation[inst1]{organization={College of Electronic Information, Sichuan University},city={Chengdu},postcode={610065},country={China}}
\affiliation[inst2]{organization={Graduate School of Interdisciplinary Information Studies, The University of Tokyo},city={Tokyo},postcode={163-8001},country={Japan}}

\begin{abstract}

For the task of low-light image enhancement, deep learning-based algorithms have demonstrated superiority and effectiveness compared to traditional methods. However, these methods, primarily based on Retinex theory, tend to overlook the noise and color distortions in input images, leading to significant noise amplification and local color distortions in enhanced results. To address these issues, we propose the Dual-Path Error Compensation (DPEC) method, designed to improve image quality under low-light conditions by preserving local texture details while restoring global image brightness without amplifying noise. DPEC incorporates precise pixel-level error estimation to capture subtle differences and an independent denoising mechanism to prevent noise amplification. We introduce the HIS-Retinex loss to guide DPEC's training, ensuring the brightness distribution of enhanced images closely aligns with real-world conditions. To balance computational speed and resource efficiency while training DPEC for a comprehensive understanding of the global context, we integrated the VMamba architecture into its backbone. Comprehensive quantitative and qualitative experimental results demonstrate that our algorithm significantly outperforms state-of-the-art methods in low-light image enhancement. The code is publicly available online at \href{https://github.com/wangshuang233/DPEC}{https://github.com/wangshuang233/DPEC}.

\end{abstract}

\begin{keyword}
Low Light Image Enhancement \sep Error Compensation \sep Vision Mamba  \sep State Space Models
\end{keyword}

\end{frontmatter}


\section{Introduction}
\label{sec1}
Low-light image enhancement has been a highly challenging task in the field of computer vision and image processing~\cite{wang2023low, WANG2023117016,liu2021benchmarking} . Images captured under low illumination conditions often exhibit dim, blurry visual effects, with severe loss of detail, significantly impacting subsequent image processing tasks such as object detection and image segmentation. In real-world detection scenarios, low-light conditions are often unavoidable. How to effectively enhance these low-light images has become a key issue in image processing research.


Traditional low-light image enhancement techniques, such as histogram equalization~\cite{abdullah2007dynamic,celik2011contextual,cheng2004simple} and gamma correction~\cite{huang2012efficient,rahman2016adaptive}, typically adjust pixel values without considering the illumination component. While these methods are effective in simple scenarios, they often degrade image quality in complex lighting conditions, leading to unnatural color shifts or loss of detail. With advancements in human visual system research, the Retinex theory~\cite{wei2018deep,yu2023two,hao2018low,cai2023retinexformer} was introduced, providing a theoretical framework for low-light image enhancement by decomposing an image into illumination and reflectance components. Many deep learning-based methods have adopted Retinex theory for low-light image enhancement, demonstrating strong performance. Among them, CNN-based approaches, such as RetinexNet~\cite{wei2018deep,zhang2019kindling,zhao2025low}, enhance visibility by estimating illumination and refining reflectance, while GAN-based approaches, such as EnlightenGAN~\cite{jiang2021enlightengan}, improve perceptual quality through adversarial training. However, due to the inherent limitations of Retinex theory, these methods often suffer from noise amplification and color distortion. In extreme low-light conditions, the instability of Retinex-based decomposition leads to significant color shifts and difficulties in effectively removing noise, thereby degrading the overall visual quality.

\begin{figure}[htb]
  \centering
  
  \begin{minipage}[t]{0.49\textwidth} 
    \centering
    \includegraphics[width=\textwidth]{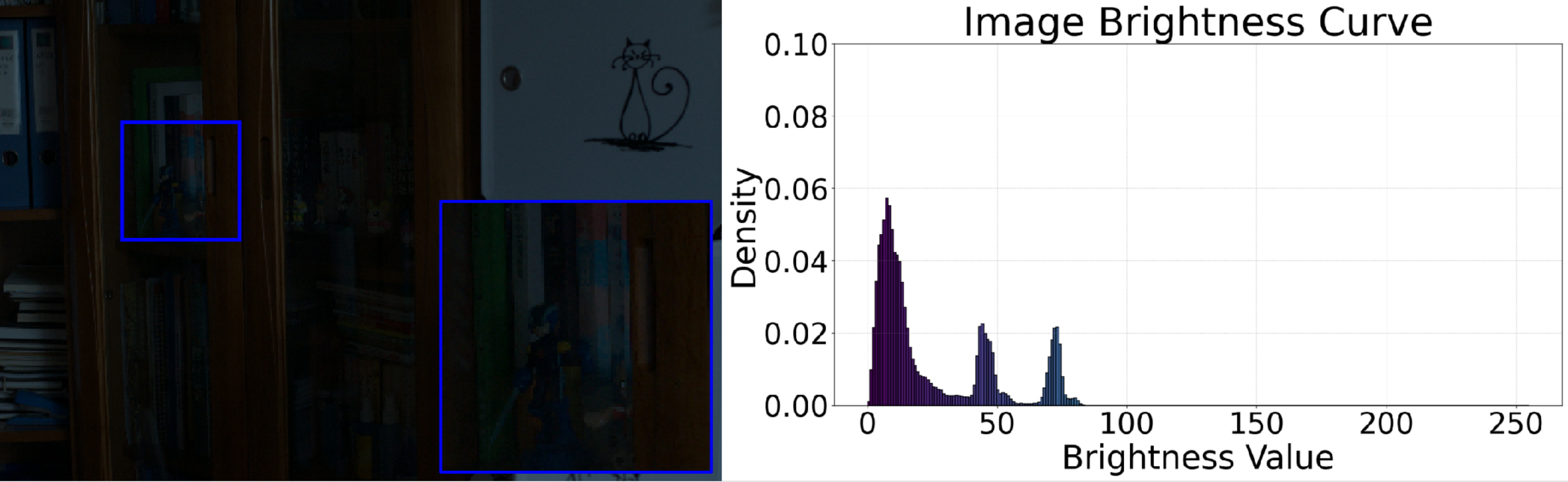}
    \caption*{(a) input}
    \label{fig:img1(a)}
  \end{minipage}\hfill
  \begin{minipage}[t]{0.49\textwidth}
    \centering
    \includegraphics[width=\textwidth]{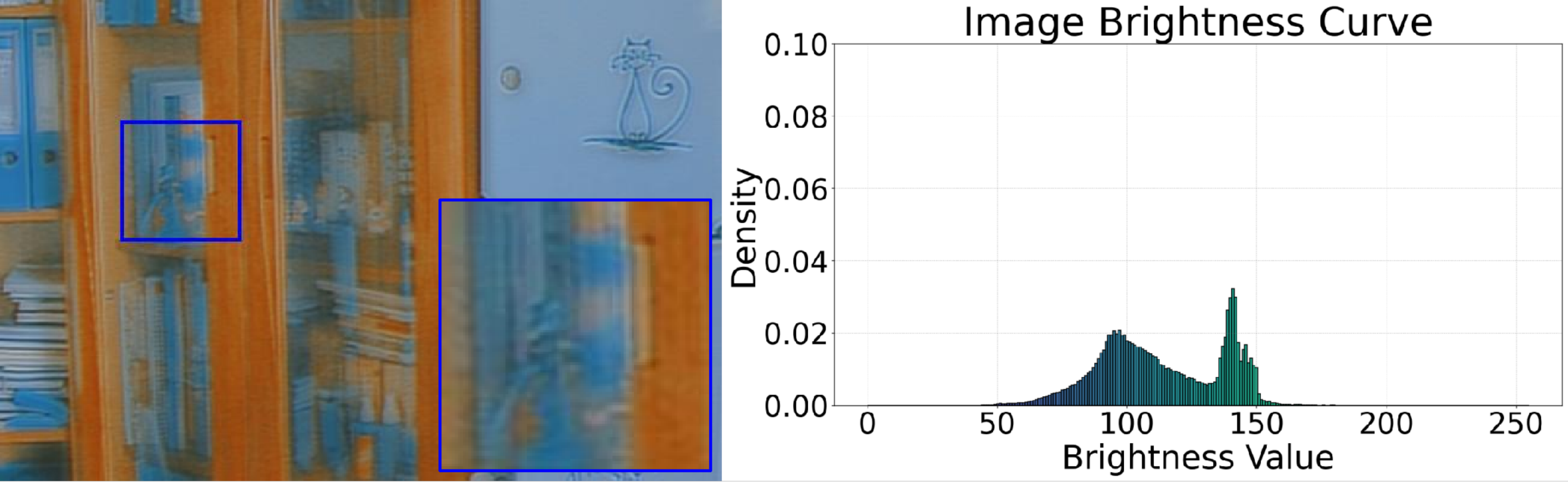}
    \caption*{(b)  RetinexNet~\cite{wei2018deep}}
    \label{fig:img1(b)}
  \end{minipage}\par\medskip

  \begin{minipage}[t]{0.49\textwidth}
    \centering
    \includegraphics[width=\textwidth]{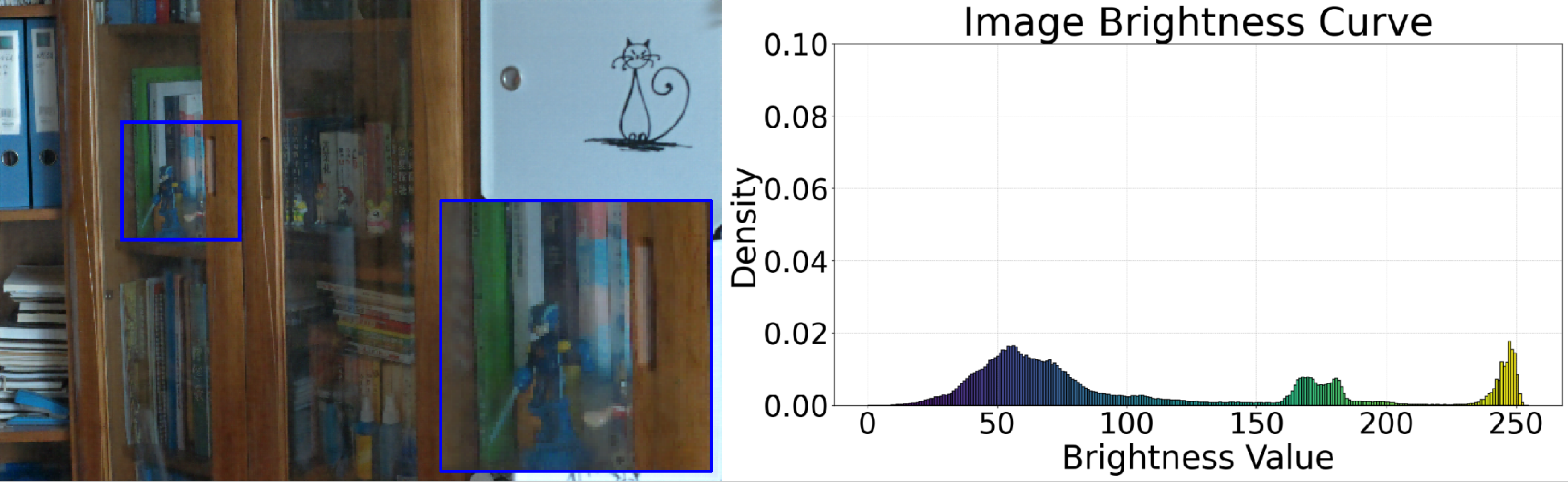}
    \caption*{(c) RetinexMamba~\cite{bai2024retinexmamba}}
    \label{fig:img1(c)}
  \end{minipage}\hfill
  \begin{minipage}[t]{0.49\textwidth}
    \centering
    \includegraphics[width=\textwidth]{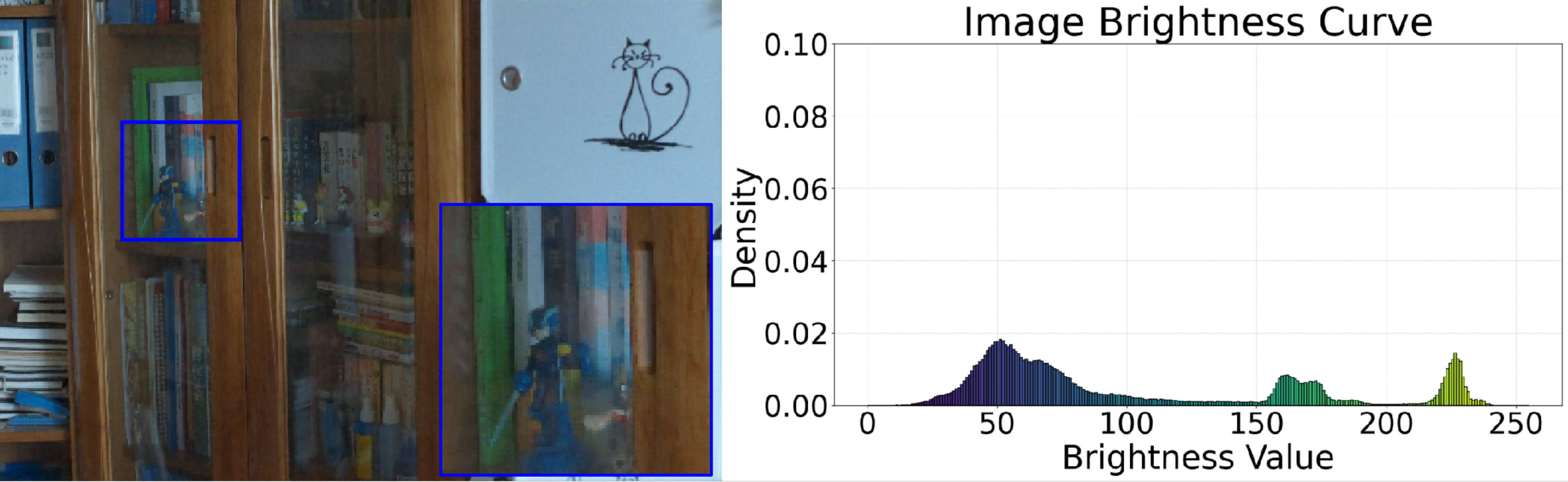}
    \caption*{(d)Retinexformer~\cite{cai2023retinexformer}}
    \label{fig:img1(d)}
  \end{minipage}\par\medskip

  \begin{minipage}[t]{0.49\textwidth}
    \centering
    \includegraphics[width=\textwidth]{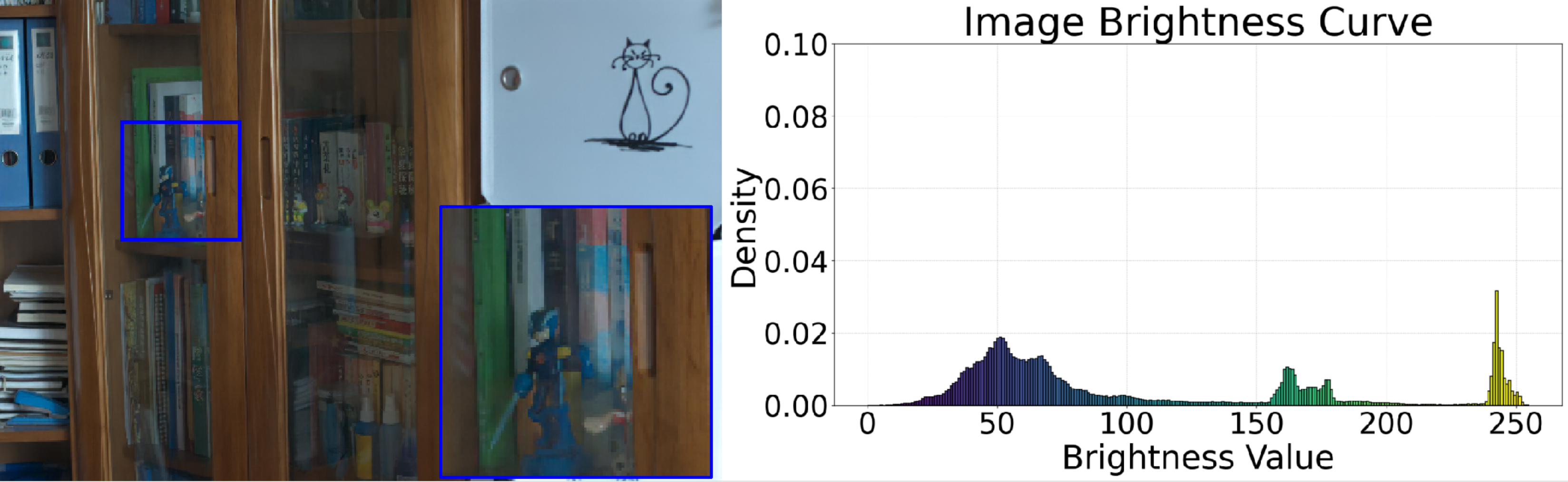}
    \caption*{(e) DPEC}
    \label{fig:img1(e)}
  \end{minipage}\hfill
  \begin{minipage}[t]{0.49\textwidth}
    \centering
    \includegraphics[width=\textwidth]{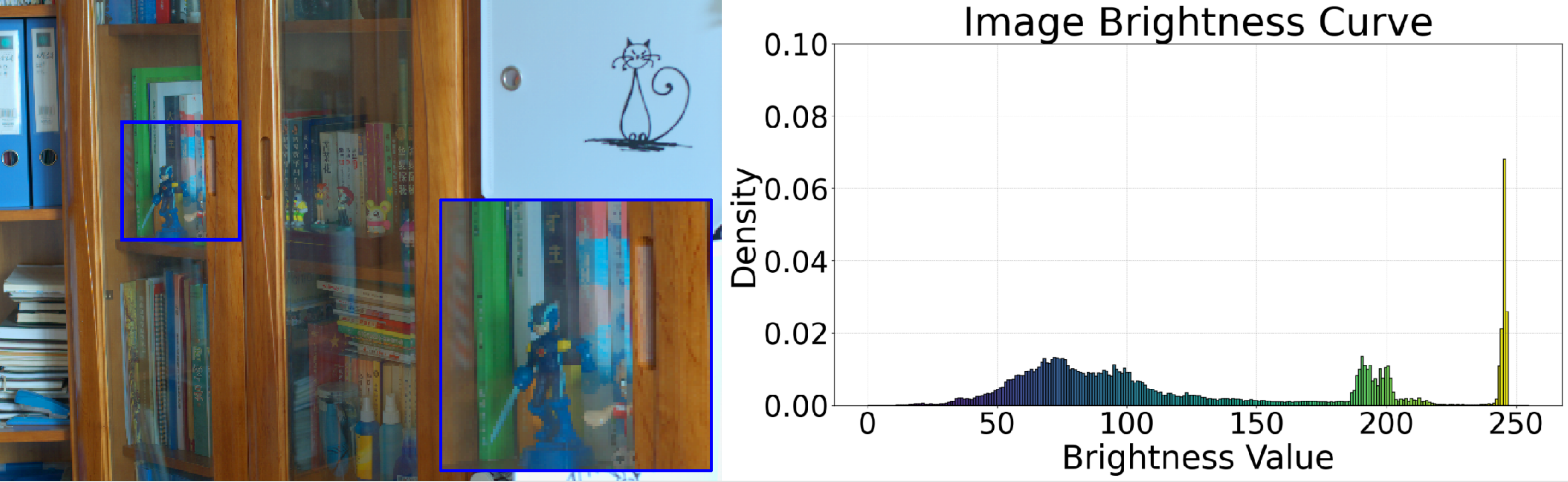}
    \caption*{(f) Ground Truth}
    \label{fig:img1(f)}
  \end{minipage}\par\medskip
  
  \caption{Visual results of different SoTA methods. Compared with other SoTA methods, our method produces results with better global brightness level and fine local textures without noise on the mainstream real-world dataset LOLv1. Please zoom in for a better look.}
  \label{fig1}
\end{figure}

Although convolution-based models, including CNNs and GANs, have been widely used in low-light image enhancement, they struggle with capturing long-range dependencies~\cite{cai2023retinexformer}. CNNs rely on local receptive fields for feature extraction, which limits their ability to model global information, resulting in inconsistent color distributions and uneven enhancement. Additionally, GANs, while effective at generating visually appealing results, often introduce artifacts, particularly when training data is insufficient, reducing their robustness. To address the limitations in long-range dependency modeling, researchers have increasingly shifted their focus toward Transformer architectures. Transformer-based methods, such as RetinexFormer~\cite{cai2023retinexformer}, Uformer~\cite{wang2022uformer}, and Restormer~\cite{zamir2022restormer}, utilize self-attention mechanisms to enhance global feature perception, thereby improving the modeling of large-scale brightness variations and structural details in low-light images. These approaches have achieved significant improvements in enhancement quality. However, Transformer models typically suffer from high computational complexity and a large number of parameters, leading to excessive resource consumption and prolonged inference times. This computational bottleneck restricts their practicality in mobile and real-time applications. Thus, a key challenge in low-light image enhancement remains balancing enhancement effectiveness with computational efficiency.

To address the aforementioned issues and limitations inherent in existing algorithms, we propose a novel low-light image enhancement algorithm, termed DPEC (Dual-Path Error Compensation). DPEC comprises two fundamental components. Firstly, we employ the Brightness Error Estimator (BEE), built on the VMamba architecture, to extract global contextual features, ensuring the preservation of local details in the enhanced images. This architecture can capture long-range semantic information from images while enhancing computational efficiency and conserving resources. The BEE accurately estimates the error component between the low-light input and a reference target, enabling the integration of this estimated error component into the original low-light input at the pixel level, thereby facilitating initial enhancement. Second, recognizing the significant noise present in the original low-light input, our algorithm incorporates a dedicated denoising stage in which the BEE is frozen, and a specifically designed DenoiseCNN (DC) is utilized to effectively eliminate noise. The final enhanced output is derived by combining the denoised result from the DC with the error component from the BEE, thereby mitigating noise amplification and color distortion throughout the enhancement process. Furthermore, to optimize the training of our algorithm and ensure enhanced performance in global brightness distribution, we introduce the HIS-Retinex loss, inspired by the Retinex theory. This loss function effectively constrains the training process, leading to improved performance in low-light image enhancement. As illustrated in  ~\Cref{fig1} (e), our algorithm demonstrates a superior alignment of global brightness distribution and local details with the ground truth when compared to Retinex-based algorithms, thereby showcasing the effectiveness of the DPEC framework.

The main contributions can be summarized as follows:
\begin{itemize}
\item We introduce the Dual-Path Error Compensation (DPEC) method, addressing common issues of noise and color distortion in low-light enhancement.
\item We propose the HIS-Retinex loss to guide the training process, ensuring better global brightness distribution.
\item We integrate the VMamba architecture to improve computational speed and resource efficiency while capturing long-range semantic information.
\item Extensive experiments demonstrate that our algorithm significantly outperforms state-of-the-art methods across six benchmarks, achieving superior image quality and efficiency.
\end{itemize}


\section{Related Works}
\label{sec2}
\subsection{Low-light Image Enhancement}
\label{sec2-1}



\textbf{Distribution Mapping Approaches:} Early low-light image enhancement techniques focused on directly adjusting pixel distributions to improve visibility. Methods like histogram equalization~\cite{abdullah2007dynamic,cheng2004simple} and gamma correction~\cite{huang2012efficient} modify intensity histograms to brighten dark areas. However, these methods operate without considering scene semantics, often leading to unnatural color shifts and exaggerated contrast, which degrade perceptual quality.

\textbf{Traditional Model-Based Approaches:} Retinex theory~\cite{fu2016fusion,guo2016lime} offers a decomposition-based strategy by separating an image into illumination and reflectance components. While this framework provides a more structured way to handle lighting inconsistencies, conventional Retinex-based methods heavily depend on handcrafted priors and heuristic tuning. As a result, they frequently introduce artifacts and suffer from poor adaptability across diverse lighting conditions. Additionally, these models tend to amplify noise during the enhancement process, further compromising image quality.

\textbf{Deep Learning-Based Approaches:} With the advent of deep learning, data-driven methods have shown remarkable effectiveness in low-light image enhancement. Retinex-inspired deep models~\cite{wei2018deep,cai2023retinexformer,zhang2019kindling,bai2024retinexmamba,zamir2022restormer,xu2022snr} integrate the decomposition principles of Retinex theory with neural networks, allowing for more robust feature extraction and adaptive enhancement. Meanwhile, Transformer-based architectures~\cite{cai2023retinexformer,wang2022uformer,zamir2022restormer} leverage self-attention to model long-range dependencies and capture global contrast variations more effectively. However, their increased computational cost and memory overhead pose challenges for real-time applications.

\subsection{Error Compensation}
\label{sec2-2}
Error compensation technology originated in the field of precision manufacturing, aiming to improve machining accuracy through dynamic correction of systematic errors.   In the 1990s, scholars such as Bryan proposed thermal error compensation theory for machine tools, which utilized temperature sensors to monitor thermal deformation errors in real time and established mathematical models to predict compensation values~\cite{bryan1990international,li2015review}.   By adjusting tool paths via feedback control, this technology significantly reduced the impact of thermal effects on machining accuracy.   Its core concept—dynamic correction based on error modeling—laid the theoretical foundation for cross-domain technology transfer.

With breakthroughs in deep learning, the concept of error compensation was introduced into the field of image processing.   In 2015, He et al. formalized the idea of residual mapping through the Residual Network (ResNet), which directly learned residual information between input and target images to address the degradation problem in deep networks~\cite{he2016deep}.   This work marked a paradigm shift in error compensation technology from physical system modeling to data-driven approaches.   Subsequent studies integrated generative adversarial networks (GANs) to propose image enhancement methods based on error compensation~\cite{ledig2017photo,salimans2016improved}.   By leveraging generators to estimate residual distributions between degraded and high-quality images and discriminators to optimize perceptual authenticity, these methods achieved remarkable improvements in tasks such as image deblurring and super-resolution. However, these GAN-based methods often suffer from high computational overhead and instability during training, particularly when dealing with long-range dependencies or complex tasks. This computational inefficiency highlights the need for alternative approaches capable of achieving high performance with lower resource consumption.

\subsection{Vision Mamba}
\label{sec2-3}

The Vision Mamba model, a modern advancement in the realm of State Space Models (SSMs), offers an efficient and scalable alternative to traditional CNN and Transformer architectures. Unlike conventional SSMs, which originate from the Kalman filter~\cite{kalman1960new} and are primarily designed for dynamic systems through differential or difference equations, Vision Mamba introduces specialized mechanisms—such as selective scanning and adaptive attention modules—that significantly enhance its suitability for visual tasks~\cite{bai2024retinexmamba,ruan2024vm,zou2024wave}. These mechanisms allow Vision Mamba to model long-range dependencies while maintaining linear computational complexity relative to input size, overcoming the quadratic cost of Transformers~\cite{zamir2022restormer,cai2023retinexformer}. This makes it particularly advantageous for high-resolution image processing, where efficiency is critical.

The selective scanning mechanism in Vision Mamba effectively captures long-range dependencies while preserving computational efficiency, a capability that has been empirically validated in tasks such as medical image segmentation~\cite{ruan2024vm} and low-level vision problems~\cite{zhu2024vision}. Its ability to maintain stability and efficiency when handling extensive visual sequences highlights its potential as a foundational model for complex image processing tasks. While Vision Mamba has demonstrated strong performance across various domains, its application to low-light image enhancement—where precise detail preservation and computational efficiency are crucial—remains an unexplored opportunity.

\begin{figure}[htb]
    \centering
    \includegraphics[width=1.0\linewidth]{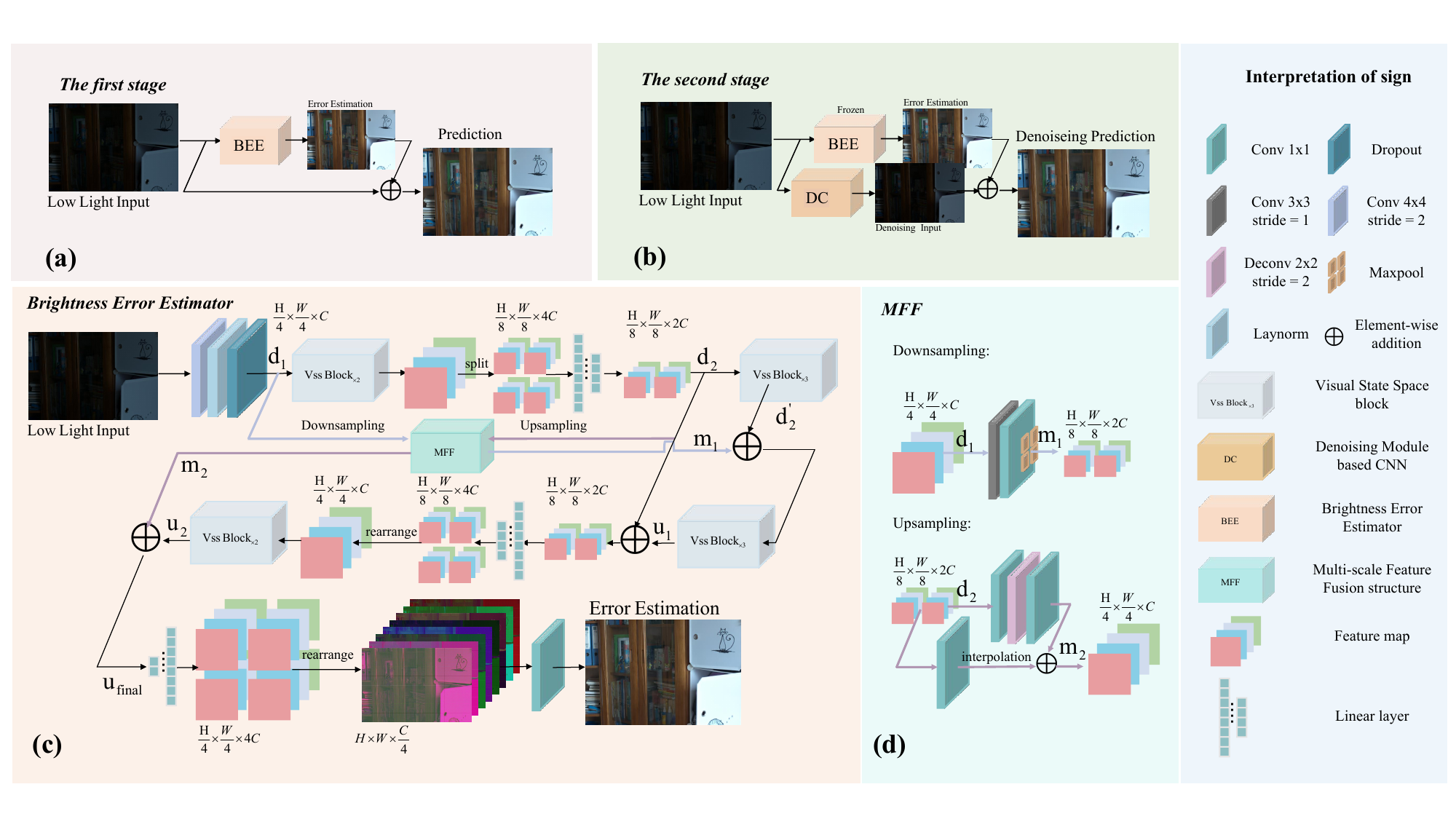}
    \caption{Overview of the Dual-Path Error Compensation Framework (DPEC).  The Dual-Path Error Compensation Framework (DPEC) is illustrated through four diagrams: (a) first-stage training (direct error estimation), (b) second-stage training (denoising and fusion), (c) the Brightness Error Estimator (BBE) architecture, and (d) its Multi-scale Feature Fusion (MFF) submodule. These diagrams clarify the framework’s dual-path strategy: (a) trains the BBE on raw low-light inputs to predict brightness errors, followed by (b) freezing the BBE and training the DenoiseCNN (DC) to denoise inputs, with their outputs fused via addition. The BBE (c) integrates multi-scale features through its MFF submodule (d), enabling cross-scale fusion for precise error estimation. By explicitly linking (a-d) to training phases and structural components, DPEC achieves robust low-light enhancement through synergistic error correction and noise removal.}
    \label{fig2}
\end{figure}


\section{Method}
\label{sec3}
Traditional Retinex-based methods for low-light image enhancement suffer from inherent limitations due to multiplicative decomposition, including noise amplification and illumination-reflectance coupling \Cref{sec3-1}. To address these challenges, a Dual-Path Error Compensation (DPEC) framework is developed, replacing error-prone decomposition with an additive error estimation paradigm. The framework integrates two core components  \Cref{sec3-2}: (1) a multi-scale Vision State Space-based Brightness Error Estimator for distortion compensation, and (2) a DenoiseCNN module combining spectral attention and dark channel processing for noise-robust enhancement. A hybrid loss function \Cref{sec3-3} combining histogram alignment, structural similarity, and perceptual constraints is further designed to guide joint optimization. Quantitative and qualitative comparisons \Cref{fig7} validate the framework's effectiveness in suppressing artifacts while preserving color fidelity and texture details.

\subsection{Preliminary}
\label{sec3-1}

Traditional Retinex-based methods decompose a low-light image \( S_{\text{Retinex}} \) into reflectance \( R \), illumination \( L \), and an inherent noise term \( N \):
\begin{equation}
   S_{\text{Retinex}} = R \cdot (L + N) = R \cdot L + R \cdot N
\end{equation}
During enhancement, the final output \( S_{\text{enhanced}} \) is obtained by adjusting \( R \) and \( L \). However, this multiplicative framework introduces two critical limitations:

\begin{itemize}
    \item \textbf{Noise Amplification:} In low-light conditions where \( L \) is small, the noise term \( N \) is inherently amplified through the product \( R \cdot N \). This results in enhanced images with severe noise and unnatural artifacts, particularly in dark regions (see ~\Cref{fig7}).
    \item \textbf{Illumination-Reflectance Coupling:} The separation of \( R \) and \( L \) is ill-posed and highly sensitive to estimation errors. Imperfect decomposition leads to color distortion (e.g., oversaturation or desaturation) and loss of texture details.
\end{itemize}

These limitations stem from the fundamental assumption of Retinex theory that illumination and reflectance can be multiplicatively decoupled. In practice, such decoupling is challenging due to the entanglement of noise, lighting variations, and scene complexity. Prior attempts to mitigate these issues (e.g., explicit noise modeling or regularization) often trade off enhancement quality for computational stability.

To address these challenges, we propose a paradigm shift through the \textbf{additive error compensation} framework in DPEC. Instead of decomposing \( S_{\text{Retinex}} \), DPEC directly estimates the error \( E \) between the low-light input \( (L + N) \) and a well-lit reference, generating the enhanced image as:

\begin{equation}
    S_{\text{DPEC}} = (L + N) + E
\end{equation}
This formulation bypasses the multiplicative noise amplification and avoids the error-prone \( R \)-\( L \) decomposition. While DPEC's technical details are elaborated in ~\Cref{sec3-2}, its core advantage lies in circumventing the intrinsic limitations of Retinex theory, as validated by the quantitative and qualitative comparisons in ~\Cref{fig7}.

\subsection{Dual-Path Error Compensation Framework}
\label{sec3-2}
Our Dual-Path Error Compensation framework (DPEC) consists of two key components: the Brightness Error Estimator (BEE) in ~\Cref{sec3-2-1} and DenoiseCNN in ~\Cref{sec3-2-2}. The BEE employs a U-shaped backbone for effective feature extraction, while DenoiseCNN focuses on reducing noise by brightening and refining the input image. The overall workflow of the algorithm can be divided into two stages (as shown in ~\Cref{fig2} (a) and ~\Cref{fig2} (b)).

\textbf{Stage 1}: The BEE serves as the core feature extraction network, accurately estimating the error component($\mathbf{E}(\mathbf{I}_{LL})$) between the low-light input and the reference target. Here, we consider the original input low-light image as $\mathbf{I}_{LL}$. This estimated error component is added pixel-wise to the original low-light input, compensating for distortions and achieving preliminary image enhancement.

\textbf{Stage 2}: To address the noticeable noise in the original low-light input, we freeze the BEE and introduce DenoiseCNN. DenoiseCNN processes the low-light input to reduce noise, producing $\mathbf{I}_{denoise}$. The denoised output is then combined with the per-pixel error estimation $\mathbf{E}(\mathbf{I}_{LL})$ from BEE to obtain the final enhanced image.

In summary, the training process of the algorithm can be represented as follows:

\begin{equation}
    \begin{aligned}
        \text{Stage 1:} \quad \mathbf{I}_{enh} &= \mathbf{I}_{LL} + \mathbf{E}(\mathbf{I}_{LL}) \\
        \text{Stage 2:} \quad \mathbf{I}_{final} &= \mathbf{I}_{denoise} + \mathbf{E}(\mathbf{I}_{LL})
    \end{aligned}
\end{equation}
\subsubsection{Brightness Error Estimator}
\label{sec3-2-1}
The overall structure of the Brightness Error Estimator (BEE) is illustrated in ~\Cref{fig2} (c). The BEE features a U-shaped feature extraction backbone based on the VMamba architecture, complemented by a Multi-scale Feature Fusion structure (MFF) for cross-scale feature integration.

For a given low-light input image, the process begins with partitioning the image into non-overlapping 4$\times$4 patches via a convolution operation performed by the Patch Embedding layer. This layer maps the channels to \( C \) (default value is 96) and normalizes the result using Layer Normalization to produce \( d_1 \). The normalized input is then fed into the U-shaped encoder for feature extraction. The encoder comprises two levels of encoding layers. At the end of the first encoding level, a patch merging operation is employed to downsample the features, reducing the spatial dimensions to \( \frac{H}{8} \times \frac{W}{8} \) and increasing the channel count to \( 2C \). Both encoding layers include 2 and 3 VSS blocks, respectively, for comprehensive deep feature extraction.

To fully utilize features of different scales, the MFF structure is applied. In ~\Cref{fig2} (d), the up and down sampling operations of MFF are represented by different colors. The low-dimensional input from the decoder is downsampled through simple convolution from top to bottom to obtain \( m_1 \), which retains more positional information and local details. Concurrently, the high-dimensional output \( d_2 \) from the first-level decoder, which contains high-level contextual information, is upsampled using deconvolution and interpolation from bottom to top to produce \( m_2 \). These outputs are then combined in the decoder section via a residual connection, enabling effective fusion of low- and high-dimensional features. The high-dimensional feature \( d_2 \) is also used through skip connections to assist in feature recovery during the upsampling phase.

Reflecting the encoder’s architecture, the decoder also consists of two levels. Prior to the first-level decoder, the output of the second-level decoder, \( d_{2}' \), is fused with \( m_1 \) to generate \( u_1 \), which is then used as the input for the first-level decoder. The first-level decoder starts with a patch expanding operation to upsample the feature map, restoring its size and reducing the channel count. The resulting \( u_1 \) is fused with \( d_2 \) at the same scale, serving as input for the second-level encoder. The decoder stages incorporate [3, 2] VSS blocks, respectively. After processing, the outputs \( u_2 \) and \( m_2 \) are fused to obtain \( u_{final} \). This final output is upsampled four times using a patch expanding operation to restore its original size and then mapped through a linear layer to adjust the number of channels, resulting in the final brightness error estimate.

\begin{figure}[htb]
    \centering
    \includegraphics[width=0.95\linewidth]{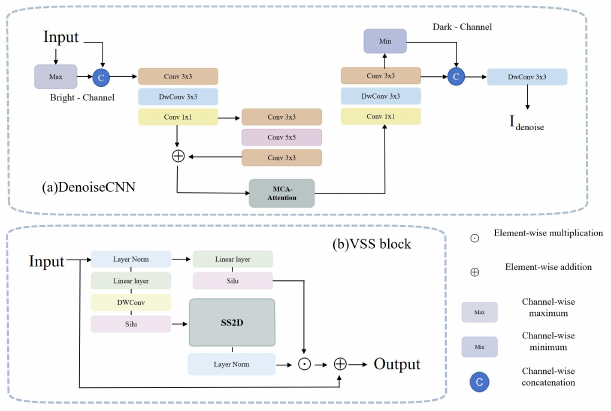}
    \caption{The structure of DenoiseCNN and the VSS block. The DenoiseCNN focuses on removing noise from low-light images through brightness enhancement and refined convolution processing to improve image clarity. The VSS block, integrated with SS2D, enhances the overall feature extraction capability of DPEC.}
    \label{fig3}
\end{figure}

\textbf{VSS Block:} The core component of the DPEC architecture is the Vision State Space (VSS) block, derived from the VMamba~\cite{zhu2024vision} framework, as shown in ~\Cref{fig3}. The input processed by this block first goes through Layer Normalization, and then it is split into two paths: the first path applies a linear layer and an activation function for feature transformation; while the second path sequentially passes through a linear layer, depthwise separable convolution, and an activation function, followed by entering the SS2D module for in-depth feature extraction. After the SS2D module, Layer Normalization is used to standardize the output features, which are then element-wise multiplied by the features output from the first path. The subsequent linear layer integrates these features, and finally, they are element-wise added with the residual connection to form the final output of the VSS block. In the implementation of the VSS block, the SILU~\cite{elfwing2018sigmoid} function is chosen by default as the activation function.
\begin{figure}[htb]
    \centering
    \includegraphics[width=0.95\linewidth]{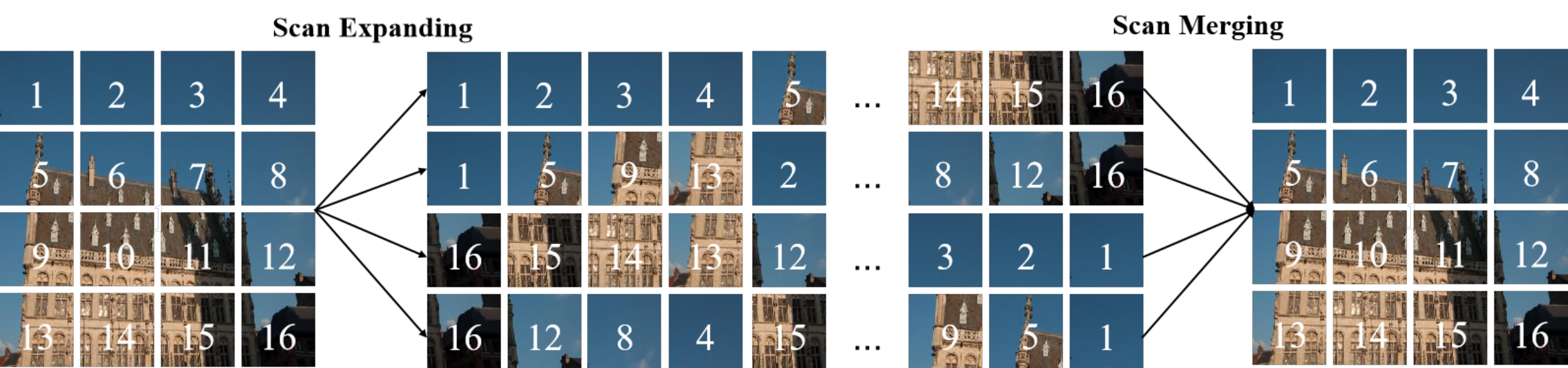}
    \caption{Scan expanding operation and Scan merging operation in
SS2D. These two operations enable the mutual transformation between feature maps and image sequences.}
    \label{fig4}
\end{figure}

\begin{algorithm}[htb]
\caption{Optimized Pseudo-code for S6 Block in SS2D}
\label{alg:S6_optimized}
\begin{algorithmic}[1]
\STATE \textbf{Input:} $x$, the feature tensor with shape $[B, L, D]$
\STATE \textbf{Parameters:} $A$, $D$ (learnable parameters)
\STATE \textbf{Operator:} \texttt{Linear(.)} (linear projection layer)
\STATE \textbf{Output:} $y$, the transformed feature tensor with shape $[B, L, D]$

\STATE Compute $\Delta, B, C \gets \texttt{Linear}(x)$
\STATE Compute temporary variable $Temp \gets \Delta \cdot A$
\STATE Update $A \gets \exp(Temp)$
\STATE Update $B \gets Temp^{-1} \cdot (\exp(Temp) - I) \cdot B$
\STATE Initialize $h_0 \gets 0$
\FOR{each time step $t$ from 1 to $L$}
    \STATE Update hidden state $h_t \gets A \cdot h_{t-1} + B \cdot x_t$
    \STATE Compute output $y_t \gets C \cdot h_t + D \cdot x_t$
\ENDFOR
\STATE Collect outputs $y \gets [y_1, y_2, \ldots, y_L]$
\STATE \textbf{return} $y$
\end{algorithmic}
\end{algorithm}

\textbf{SS2D}: The SS2D structure mainly consists of three parts: scan expansion module, S6 feature extraction module, and scan merging module, as shown in ~\Cref{fig4}. The scan expansion module performs sequential expansion of the input image along four different directions (top-left to bottom-right, bottom-right to top-left, top-right to bottom-left, and bottom-left to top-right). This ensures a thorough scan of the image in all directions, helping to capture multi-directional features. Subsequently, the S6 module performs feature extraction on these sequences obtained from different directions based on SSM. The SSM pseudocode~\cite{ruan2024vm} is provided here. Finally, the scan merging module sums and merges these feature sequences from different directions, restoring the processed feature map to the original spatial dimensions of the input image. The S6 block incorporates a selective mechanism based on the S4 block, enhancing the model's ability to discriminate and filter feature information, effectively retaining valuable information and discarding redundant parts.

\subsubsection{DenoiseCNN}  
\label{sec3-2-2}  

In practical low-light image enhancement, traditional methods often directly multiply or add the input image with the estimated illumination, neglecting the inherent noise component. To address this limitation, we design DenoiseCNN, a dedicated deep network module for noise removal in low-light images. DenoiseCNN processes the low-light input ($\mathbf{I}_{LL}$) to obtain the denoised result ($\mathbf{I}_{denoise}$).

As shown in ~\Cref{fig3}, DenoiseCNN begins with a brightening operation to capture the overall structural information of the input image. The brightened image is then merged with the original input and passed through a preprocessing convolutional layer~\cite{liu2021learning}, which extracts local detail features while expanding the number of feature channels to enrich the representation. The output of this stage is preserved as a residual connection and further refined through stacked convolutional layers.  

A key innovation of DenoiseCNN is its integration of the Multi-Spectral Channel Attention (MCA) mechanism~\cite{qin2021fcanet}, which systematically enhances high-frequency details (e.g., edges and textures) while suppressing noise through a frequency-aware framework. First, input features are decomposed into distinct frequency bands using the Discrete Cosine Transform (DCT)~\cite{ahmed1974discrete}, explicitly isolating low-frequency components (smooth regions) from high-frequency components (noise-sensitive edges and textures). Next, a channel attention mechanism assigns adaptive weights to each frequency band based on its global context, prioritizing features most critical for detail recovery. By learning cross-channel spectral dependencies, the MCA dynamically strengthens informative high-frequency bands and suppresses noise-contaminated components. Finally, the enhanced frequency features are reconstructed into the spatial domain using the inverse DCT and fused with residual features from the pre-processing stage, ensuring balanced preservation of structural integrity and fine details.

The MCA mechanism provides several distinct advantages in the denoising process. First, it explicitly focuses on high-frequency components that are critical for visual quality, ensuring better recovery of edges and textures. Second, it adaptively suppresses noise without compromising essential image details, thereby preserving naturalness and structural clarity. Finally, the frequency-domain processing allows for a more precise separation of noise and details, enhancing the overall robustness of the enhancement process.  

Additionally, inspired by the Dark Channel Prior, we apply a dark channel processing step to further refine the naturalness and structural clarity of the denoised image. This comprehensive approach enables DenoiseCNN to achieve superior performance in both noise suppression and detail preservation, making it a robust solution for low-light image enhancement.

\subsection{Loss Function}
\label{sec3-3}
By carefully designing the loss function, we can effectively guide the model training process and improve its performance. In the task of low-light image enhancement, relying on a single loss function is difficult to achieve satisfactory results. Therefore, inspired by Lin's work, we designed and configured multiple loss functions in the model training phase to promote model learning and obtain high-quality enhanced output. Specifically, we utilized traditional ssim~\cite{wang2004image} loss, perceptual loss~\cite{johnson2016perceptual}, inner loss, smooth loss~\cite{bontonou2019introducing}, tv loss~\cite{lin2023smnet} and HIS-Retinex loss as components of the total loss function.

\textbf{HIS-Retinex Loss:} Retinex theory decomposes an image into illumination \(L\) and reflectance \(R\) components, where the reflectance represents the intrinsic color and texture of the scene. Traditional Retinex-based methods often focus on estimating the illumination component but may neglect the accurate recovery of reflectance, leading to color distortion and unnatural artifacts. To address this, we introduce the HIS-Retinex Loss, which explicitly constrains the reflectance distribution of the enhanced images.

The HIS-Retinex Loss consists of two components: \textbf{Histogram Loss} and \textbf{Reflectance Histogram Loss}. The Histogram Loss quantifies the discrepancy in pixel value distribution between the predicted output and the target image, ensuring that the global brightness distribution aligns with real-world conditions. The Reflectance Histogram Loss focuses on the reflectance component, which is crucial for preserving color accuracy and texture details.
\begin{equation}
    R_{target} = \frac{I_{target}}{L_{target}}, \quad R_{pred} = \frac{I_{pred}}{I_{LL}}
\end{equation}
where \(L_{target}\) is the estimated illumination of the target image, and \(I_{pred}\) is the predicted enhanced image. The Reflectance Histogram Loss is then computed as the difference between the histograms of \(R_{target}\) and \(R_{pred}\):
\begin{equation}
    \mathcal{L}_{{histo}}=\frac{1}{2}\sum_{i=0}^{255}\left|\frac{H_{pred}\left(i\right)}{\sum_{i=0}^{255} H_{pred}\left(i\right)}-\frac{H_{target}\left(i\right)}{\sum_{i=0}^{255} H_{target}\left(i\right)}\right|+\frac{1}{2} \sum_{i=0}^{255} \left| \frac{H_{R_{pred}}(i)}{\sum_{i=0}^{255} H_{R_{pred}}(i)} - \frac{H_{R_{target}}(i)}{\sum_{i=0}^{255} H_{R_{target}}(i)} \right|    
\end{equation}
Where $H_{pred}$ and $H_{target}$ represent the histogram distributions of the predicted result and the reference target, respectively.

\textbf{Ssim Loss: }In low-light image enhancement, structural distortions such as blurring often degrade the quality of the output images. To address this issue and ensure structural consistency, we employ the Structural SIMilarity (SSIM) loss~\cite{wang2004image}, which measures the structural similarity between the input and output images. The SSIM loss is defined as:

\begin{equation}
\mathcal{L}_{ssim} = 1 - \left( \frac{2\mu_x \mu_y + C_1}{\mu_x^2 + \mu_y^2 + C_1} \cdot \frac{2\sigma_{xy} + C_2}{\sigma_x^2 + \sigma_y^2 + C_2} \right)
\end{equation}

where \(\mu_x\) and \(\mu_y\) represent the mean intensity values of the input and output images, \(\sigma_x^2\) and \(\sigma_y^2\) denote their variances, \(\sigma_{xy}\) is the covariance between the input and output images, and \(C_1\) and \(C_2\) are small constants introduced to stabilize the division and avoid division by zero. By minimizing the SSIM loss, our model ensures that the enhanced images retain structural fidelity, which is critical for maintaining visual quality in low-light enhancement tasks.

\textbf{Perceptual Loss: } To further enhance the visual quality of the enhanced images, we incorporate perceptual loss~\cite{johnson2016perceptual}, which leverages high-level feature representations extracted from a pre-trained VGG network. This loss function measures the discrepancy between the feature maps of the predicted image (\(I_{pred}\)) and the ground truth image (\(I_{gt}\)), ensuring that the output images are visually appealing and semantically consistent. The perceptual loss is defined as:

\begin{equation}
\mathcal{L}_{Perceptual} = \| \phi_{ij} (I_{\text{pred}}) - \phi_{ij} (I_{\text{gt}}) \|_1,
\end{equation}

where \(\phi_{ij}\) denotes the feature maps obtained from the \(j\)-th convolutional layer of the \(i\)-th block in the VGG-16 network, and \(\|\cdot\|_1\) represents the \(L_1\) norm, which measures the absolute difference between the feature maps. By minimizing this loss, our model bridges the gap between low-level pixel-wise differences and high-level semantic understanding, significantly improving the visual quality of the enhanced images.

\textbf{Tv Loss: } To mitigate the undesirable noise amplification that often occurs during the brightness enhancement process, we employ Total Variation (TV) loss~\cite{lin2023smnet}. Noisy images typically exhibit high variance, and the TV loss addresses this issue by minimizing the total changes between adjacent pixels, thereby promoting smoother and cleaner outputs. The TV loss is defined as:

\begin{equation}
\mathcal{L}_{tv}(I_{\text{pred}}) = \sum_{i=1}^{H} \sum_{j=1}^{W} \sqrt{(I_{\text{pred}}(i,j) - I_{\text{pred}}(i+1,j))^2 + (I_{\text{pred}}(i,j) - I_{\text{pred}}(i,j+1))^2},
\end{equation}

where \(I_{pred}(i,j)\) denotes the pixel value at position \((i,j)\) in the predicted image \(I_{pred}\), and \(H\) and \(W\) represent the height and width of the image, respectively. By incorporating the TV loss, our model effectively reduces noise while preserving essential structural details, leading to visually cleaner and more natural results.

\textbf{Smooth L1 Loss: }To balance the trade-off between robustness and sensitivity to outliers, we adopt the Smooth L1 Loss~\cite{bontonou2019introducing} for our optimization objective. Unlike the standard L1 loss, which is highly sensitive to outliers, or the L2 loss, which may lead to overly smooth results, the Smooth L1 Loss combines the advantages of both. It behaves like an L2 loss for small errors (ensuring smoothness) and like an L1 loss for large errors (ensuring robustness). The Smooth L1 Loss is defined as:

\begin{equation}
\mathcal{L}_{\text{smooth}}(x, y) = 
\begin{cases} 
\frac{1}{2}(x - y)^2 & \text{if } |x - y| < 1, \\
|x - y| - \frac{1}{2} & \text{otherwise},
\end{cases}
\end{equation}

where \(x\) and \(y\) represent the predicted and target values, respectively. By minimizing the Smooth L1 Loss, our model achieves a balance between preserving fine details and handling outliers, leading to more stable and visually consistent results.

\textbf{Inner Loss: }To measure the alignment between the predicted image and the target low-light image, we introduce the Inner Loss, which computes the normalized dot product between the flattened versions of the predicted image (\(pred_t\)) and the target low-light image (\(LL_t\)). The Inner Loss is defined as:

\begin{equation}
\mathcal{L}_{\text{inner}} = \frac{\langle pred\_t\_flatten, LL\_t\_flatten \rangle}{N \times H \times W \times C},
\end{equation}

where \(\langle \cdot, \cdot \rangle\) denotes the dot product between the flattened vectors of \(pred_t\) and \(LL_t\), and \(N\), \(H\), \(W\), and \(C\) represent the batch size, height, width, and number of channels of the target image \(LL_t\), respectively. This loss quantifies the similarity between the predicted and target images in a high-dimensional space, normalized by the total number of elements in the image tensor. By minimizing the Inner Loss, our model encourages the predicted image to align closely with the target low-light image, ensuring that the enhanced results maintain fidelity to the original scene while preserving structural and semantic consistency.

\textbf{Total Loss:}The total loss function is set as follows:
\begin{equation}
    \mathcal{L}_{total} =w_s\mathcal{L}_{ssim} + w_p\mathcal{L}_{perceptual} + w_{inner}\mathcal{L}_{inner} + w_h\mathcal{L}_{{histo}} + w_{tv}\mathcal{L}_{{tv}} +w_{smooth}\mathcal{L}_{{smooth}}
\end{equation}
Building on the findings of Lin et al.~\cite{lin2023smnet} and our own experimental observations, we determined the hyperparameters $w_s$, $w_p$, $w_{inner}$, $w_h$, $w_{tv}$, and $w_{smooth}$ to be 2.0, 1.2, 1.0, 1.0, 0.01, and 0.8, respectively. Detailed fusion experiments supporting these parameter choices are presented in ~\Cref{sec4-4}.
\label{sec4}
\section{Experiments}
\label{sec4-1}
\subsection{Datasets and Implementation Details}
To conduct a systematic and comprehensive evaluation of DPEC, we trained, validated, and tested the model's performance on the LOL series datasets and the LSRW dataset. We also implemented appropriate data augmentation and optimization techniques to maximize the model's performance. 

To validate the issues of noise amplification and color distortion inherent in the Retinex theory, we employed BEE to estimate the reflectance ($ \mathbf{R}(\mathbf{I}_{LL})$) in the Retinex model. Guided by the Retinex theory, we performed low-light image restoration, and simultaneously applied DenoiseCNN to denoise the low-light input($\mathbf{L}_{denoise}$). This method is termed DPEC-Retinex. The method for obtaining DPEC-Retinex can be referred to in \cref{sec3-1}, where the DPEC method is described. The process of obtaining DPEC-Retinex can be referred to in \Cref{fig8}. It can be expressed by the following equation:
\begin{equation}
    \begin{aligned}
        \text{Stage 1:} \quad \mathbf{I}_{enh} &= \mathbf{L}_{LL} * \mathbf{R}(\mathbf{I}_{LL}) \\
        \text{Stage 2:} \quad \mathbf{I}_{final} &= \mathbf{L}_{denoise} * \mathbf{R}(\mathbf{I}_{LL})
    \end{aligned}
\end{equation}

\textbf{LOL:} The LOL dataset consists of two versions~\cite{yang2021sparse,zhang2021beyond}: v1, which contains 485 images for training, 100 images for validation, and 15 images for testing, and v2, which has two subsets. The LOLv2 real subset includes 689 pairs for training, 100 pairs for validation, and 100 pairs for testing, all captured in real scenes. The LOLv2 synthetic subset contains 1,000 synthesized pairs, divided into 800 pairs for training, 100 pairs for validation, and 100 pairs for testing. To augment the limited training data, we applied vertical and horizontal flips during the data preprocessing stage.

\textbf{LSRW:} The LSRW dataset~\cite{jiang2023low} comprises 5,650 image pairs captured in various scenes. We utilized 5,000 pairs for training, 600 pairs for validation, and the remaining 50 pairs for testing. The large size of this dataset ensures that the model can generalize well across a variety of low-light conditions.

\textbf{Implementation Details:} The DPEC model was implemented using the PyTorch framework and trained on a Linux system equipped with an NVIDIA RTX 4090 GPU, CUDA 12.4, Python 3.10, and PyTorch 2.1. The training procedure consists of two distinct stages, each designed to tackle different aspects of low-light image enhancement.

In the first stage, we trained the Brightness Error Estimator (BEE) independently using the training data, while validating the model on a separate validation set. The output of this stage is generated by adding the illumination estimate from the BEE to the low-light input. The initial learning rate was set to \(5 \times 10^{-4}\), and it was decayed using a cosine annealing schedule to a minimum of \(5 \times 10^{-5}\) over 600 epochs. Validation loss was monitored during this process to avoid overfitting.

In the second stage, we froze the parameters of the BEE and introduced the DenoiseCNN for noise reduction on the low-light input. The output of the DenoiseCNN replaced the initial low-light input, producing the final enhanced image. The initial learning rate was set to \(2 \times 10^{-3}\), decayed using the cosine annealing method to a minimum of \(2 \times 10^{-4}\) over another 600 epochs. The validation set was also employed to fine-tune the model and prevent overfitting.

During testing, the trained model was evaluated on the respective test sets of the LOL and LSRW datasets. The Adam optimizer was used throughout both stages, with momentum terms \( \beta_1 = 0.9 \) and \( \beta_2 = 0.999 \) to iteratively optimize the model parameters.

\textbf{Metrics:} We adopted average image quality metrics including PSNR, SSIM, and LPIPS~\cite{zhang2018unreasonable}. PSNR (Peak Signal-to-Noise Ratio) is used to measure the visual error between two images, with higher values indicating better enhancement effects by the algorithm. SSIM (Structural Similarity Index) estimates the similarity between two images, with higher values indicating better preservation of high-frequency details and structures by the algorithm. LPIPS (Learned Perceptual Image Patch Similarity) assesses the perceptual difference between the algorithm-generated image and the real image, with lower values indicating that the generated image aligns more closely with human visual perception.
\label{sec4-2}
\subsection{Low-light Image Enhancement}

\begin{figure}[htbp]
    \centering
    \includegraphics[width=0.8\linewidth]{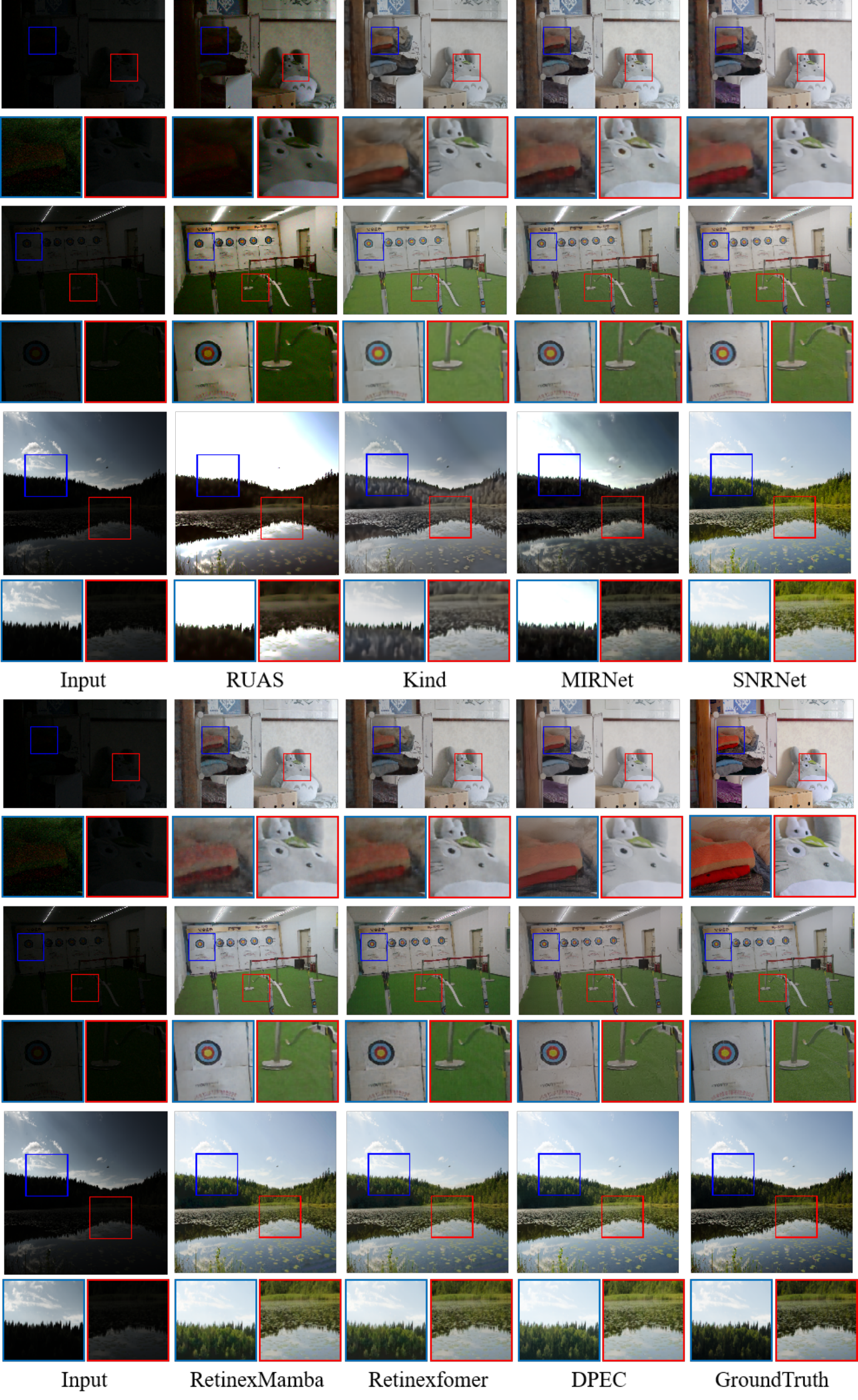}
    \caption{Results on LOL-v1 (top), LOL-v2-real (middle) and LOL-v2-synthetic (bottom). Our method effectively enhances the visibility and preserves the color.}
    \label{fig5}
\end{figure}
\begin{figure}[htbp]
    \centering
    \includegraphics[width=1.0\linewidth]{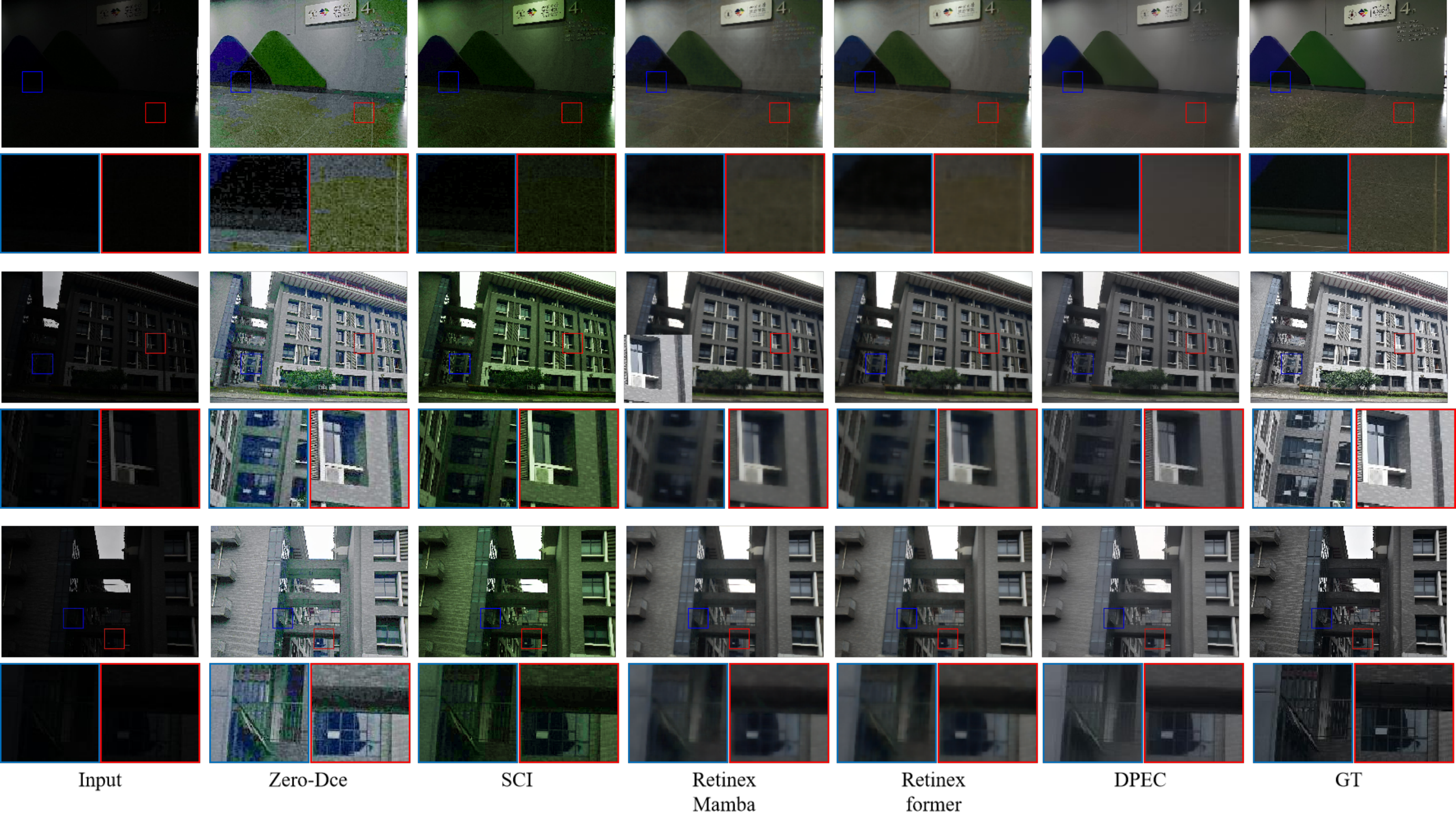}
    \caption{Results on LSRW. Other algorithms either generate over-exposed and noisy images,
or introduce black spot corruptions and unnatural artifacts. While DPEC can restore well-exposed structural contents and textures.}
    \label{fig6}
\end{figure}

\begin{table}[htbp]
\centering
\caption{Performance metrics compared for different methods on LOLV1 and LOLV2 datasets (real and synthetic). The highest results are in \textcolor{red}{red} color while the second highest result is in \textcolor{blue}{blue} color.}
\resizebox{0.95\textwidth}{!}{%
\begin{tabular}{lcccccccc}
\toprule
& \multicolumn{2}{c}{Complexity} & \multicolumn{2}{c}{LOLV1} & \multicolumn{2}{c}{LOLV2-real} & \multicolumn{2}{c}{LOLV2-synthetic} \\
\cmidrule(r){2-3}\cmidrule(r){4-5} \cmidrule(r){6-7} \cmidrule(r){8-9}
Method &FLOPs(G) &Params(M) & PSNR$\uparrow$ & SSIM$\uparrow$   & PSNR$\uparrow$  & SSIM$\uparrow$   & PSNR$\uparrow$  & SSIM$\uparrow$   \\
\midrule
SID~\cite{chen2019seeing}     &13.73 &7.76 &14.35 &0.436 &13.24 &0.442 &15.04 &0.610\\
3DLUT~\cite{zeng2020learning}   &0.075 &0.59 &14.35 &0.445 &17.59 &0.721 &18.04 &0.800\\
DeepUPE~\cite{wang2019underexposed} &21.10 &1.02 &14.38 &0.446 &13.27 &0.452 &15.08 &0.623\\
RF~\cite{kosugi2020unpaired}      &46.23 &21.54&15.23 &0.452 &14.05 &0.458 &15.97 &0.632\\
DeepLPF~\cite{moran2020deeplpf} &5.86  &1.77 &15.28 &0.473 &14.10 &0.480 &16.02 &0.587\\
IPT~\cite{chen2021pre}     &6887  &115.31 &16.27 &0.504 &19.80 &0.813 &18.30 &0.811\\
UFormer~\cite{wang2022uformer} &12.00 &5.29 &16.36 &0.771 &18.82 &0.771 &19.66 &0.871\\
RetinexNet~\cite{wei2018deep} &587.47 &0.84 &16.77 &0.560 &15.47 &0.567 &17.13 &0.798\\
Sparse~\cite{yang2021sparse}  &53.26 &2.33 &17.20 &0.640 &20.06 &0.816 &22.05 &0.905\\
EnGAN~\cite{jiang2021enlightengan}   &61.01 &114.35 &17.48 &0.650 &18.23 &0.617 &16.57 &0.734\\
RUAS~\cite{liu2021retinex}    &0.83 &0.003 &18.23 &0.720 &18.37 &0.723 &16.55 &0.652\\
FIDE~\cite{xu2020learning}    &28.51 &8.62 &18.27 &0.665 &16.85 &0.678 &15.20 &0.612\\
DRBN~\cite{yang2021band}    &48.61 &5.27 &20.13 &0.830 &20.29 &0.831 &23.22 &0.927\\
KinD~\cite{zhang2019kindling}    &34.99 &8.02 &20.86 &0.790 &14.74 &0.641 &13.29 &0.578\\
Restormer~\cite{zamir2022restormer} &144.25 &26.13 &22.43 &0.823 &19.94 &0.827 &21.41 &0.830\\
MIRNet~\cite{zamir2020learning}  &785 &31.76 &24.14 &0.830 &20.02 &0.820 &21.94 &0.876\\
SNR-Net~\cite{xu2022snr} &26.35 &4.01 &24.61 &0.842 &21.48 &\textcolor{blue}{0.849} &24.14 &0.928\\
RetinexMamba~\cite{bai2024retinexmamba} &34.75 &24.1 &24.03 &0.827  &22.00 &\textcolor{blue}{0.849} &\textcolor{blue}{25.89} &\textcolor{red}{0.943}\\
Retinexformer~\cite{cai2023retinexformer} &15.57 &1.61 &\textcolor{red}{25.16} &\textcolor{blue}{0.845} &\textcolor{blue}{22.80} &0.840 &25.67 &0.930\\
\midrule
DPEC-Retinex  &16.11 &2.58 &20.70 &0.783 &22.02 &0.797  &25.11  &0.938\\
\textbf{DPEC} &16.11 &2.58 &\textcolor{blue}{24.80} &\textcolor{red}{0.855} &\textcolor{red}{22.89} &\textcolor{red}{0.863} &\textcolor{red}{26.19} &\textcolor{blue}{0.939}\\
\bottomrule
\end{tabular}%
}
\label{table1}
\end{table}

\begin{table}[htbp]
\centering
\caption{Performance metrics compared for different methods on LSRW datasets. The highest results are in \textcolor{red}{red} color while the second highest result is in \textcolor{blue}{blue} color.}
\resizebox{0.8\textwidth}{!}{%
\begin{tabular}{lcccccc}
\toprule
Method & LIME~\cite{guo2016lime}  &RetinexNet~\cite{wei2018deep} & DRBN~\cite{yang2021band}  &Zero-DCE~\cite{guo2020zero} &MIRNet~\cite{zamir2020learning} &RUAS~\cite{liu2021retinex}  \\
\midrule
PSNR$\uparrow$   &17.342 &15.609 &16.734 &15.867 &16.470 &14.271 \\
SSIM$\uparrow$   &0.520  &0.414  &0.507  &0.443  &0.477  &0.461  \\
LPIPS$\downarrow$&0.471  &0.454  &0.457  &0.411  &0.430  &0.501  \\
\end{tabular}%
}
\resizebox{0.8\textwidth}{!}{%
\begin{tabular}{lcccccc}
\toprule
Method & SNRNet~\cite{xu2022snr} &Uformer~\cite{wang2022uformer} &RetinexMamba~\cite{bai2024retinexmamba} &Retinexformer~\cite{cai2023retinexformer}  & \textbf{DPEC}\\
\midrule
PSNR$\uparrow$   &16.499 &16.591 &19.536  &\textcolor{blue}{19.570}      &\textcolor{red}{19.643}\\
SSIM$\uparrow$   &0.505  &0.494  &0.576   &\textcolor{red}{0.578}   &\textcolor{blue}{0.576}\\
LPIPS$\downarrow$&0.419  &0.435  &0.412  &\textcolor{blue}{0.403}   &\textcolor{red}{0.360}\\
\bottomrule
\end{tabular}%
}
\label{table2}
\end{table}
\begin{figure}[htb]
    \centering
    \includegraphics[width=0.95\linewidth]{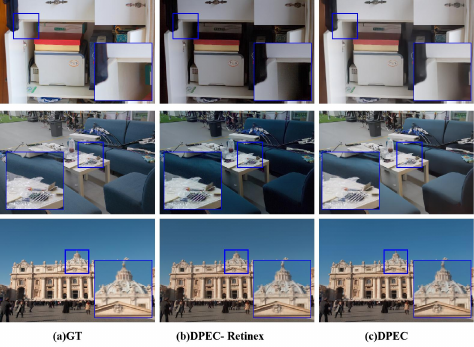}
    \caption{Comparison of local details between DPEC and DPEC-Retinex. Referring to the local details provided by the ground truth labels, DPEC-Retinex exhibits greater noise, more significant color distortion, and a higher degree of detail loss in the darker regions compared to DPEC.}
    \label{fig7}
\end{figure}
\begin{figure}[!h]
    \centering
    \includegraphics[width=0.95\linewidth]{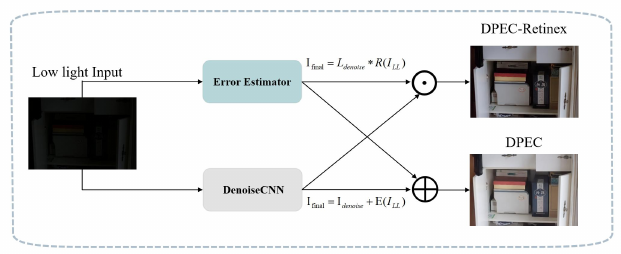}
    \caption{The workflow diagrams of DPEC and DPEC-Retinex are illustrated. DPEC-Retinex utilizes the BEE to estimate the reflectance component and, following the Retinex theory, multiplies the reflectance with the illumination component to produce the final output}.
    \label{fig8}
\end{figure}

\textbf{Quantitative Results:} In ~\Cref{table1}, ~\Cref{table2}, and ~\Cref{table3}, We evaluate DPEC against state-of-the-art (SOTA) methods on the LOL and LSRW datasets using PSNR, SSIM, and LPIPS metrics. As shown in ~\Cref{table1,table2}, DPEC consistently outperforms existing methods across all benchmarks. On the LOLv1 dataset, DPEC achieves a PSNR of 24.80 dB and SSIM of 0.855, surpassing RetinexFormer by 0.36 dB in PSNR and 1\% in SSIM. For LOLv2-real and LOLv2-synthetic, DPEC improves PSNR by 0.09 dB and 0.65 dB, respectively, while maintaining competitive SSIM scores. On the LSRW dataset, DPEC achieves a PSNR of 19.643 dB and LPIPS of 0.360, outperforming RetinexFormer by 0.073 dB in PSNR and 0.043 in LPIPS.
\begin{figure}
    \centering
    \includegraphics[width=0.95\linewidth]{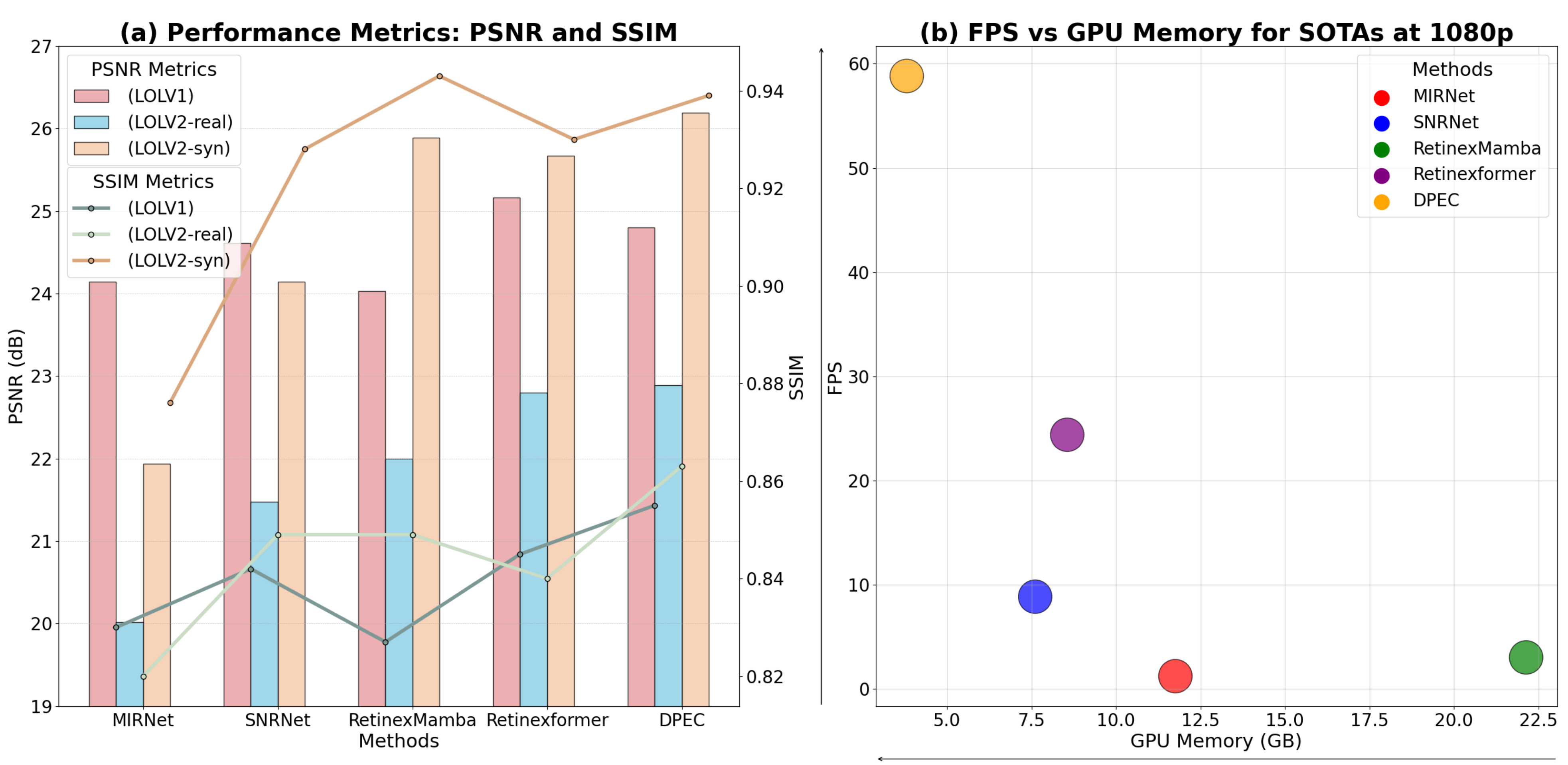}
    \caption{DPEC vs. SOTAs: Image Enhancement Quality and Inference Efficiency Comparison. Comparison between DPEC and SOTAs regarding image enhancement quality (a), as well as inference resource consumption and real-time inference capabilities (b). DPEC demonstrates a significant advantage over these algorithms in terms of image enhancement effectiveness, resource efficiency, and real-time inference performance.}
    \label{fig9}
\end{figure}
In terms of computational efficiency, as shown  ~\Cref{table3}, in DPEC demonstrates significant advantages. At 1080P resolution, DPEC achieves a 2.5$\times$ faster inference speed than RetinexFormer while consuming only 44\% of the memory resources. Compared to RetinexMamba, DPEC is 20$\times$ faster and uses 5.8$\times$ less memory, highlighting its suitability for real-time applications.

DPEC also outperforms traditional Retinex-based methods (RetinexNet~\cite{wei2018deep}, KinD~\cite{zhang2019kindling}) and Transformer-based approaches (Uformer~\cite{wang2022uformer}, Restormer~\cite{zamir2022restormer}) in both PSNR and SSIM metrics. For instance, DPEC achieves a PSNR improvement of 3.94 dB over RetinexNet and 2.37 dB over Uformer on the LOLv1 dataset. These results demonstrate DPEC’s superior ability to handle noise and color distortion while preserving image details.

Compared to the DPEC-Retinex algorithm based on the Retinex theory, the DPEC approach achieves more favorable enhancement quality by directly compensating for errors in low-light images within the LOL dataset. Specifically, for the DPEC-Retinex method, DPEC outperformed on the LOLv1 dataset, achieving a lead of 4.10 dB in PSNR and 0.072 in SSIM. On the LOLv2-real and LOLv2-syn datasets, DPEC demonstrated advantages of 0.87 dB and 1.08 dB in PSNR, as well as 0.066 and 0.001 in SSIM, respectively.
\begin{table}[!h]
\centering
\caption{Comparison of Inference Time and GPU Memory Usage for Different SOTAs. The average time (seconds) and GPU memory (G) costs of different methods consumed on the image with the size of 1920$\times$1080 (1080P), 2560$\times$1440 (2K) and 3840$\times$2160 (4k) respectively, during inference. OOM denotes the out-of-memory error and '/' denotes unavailable.}
\resizebox{0.95\textwidth}{!}{%
\begin{tabular}{lcccccc}
\toprule
& \multicolumn{2}{c}{1920$\times$1080} & \multicolumn{2}{c}{2560$\times$1440} & \multicolumn{2}{c}{3840$\times$2160} \\
\cmidrule(lr){2-3}\cmidrule(lr){4-5}\cmidrule(lr){6-7}
Method & Inference time/s $\downarrow$ & GPU Mem/G $\downarrow$ & Inference time/s $\downarrow$ & GPU Mem/G $\downarrow$ & Inference time/s $\downarrow$ & GPU Mem/G $\downarrow$ \\
\midrule
MIRNet & 0.819 & 11.76 & 1.528 & 20.40 & / & OOM \\ 
SNRNet & 0.113 & 7.61 & 0.352 & 21.45 & / & OOM \\
RetinexMamba & 0.331 & 22.14 & / & OOM & / & OOM \\
Retinexformer & 0.041 & 8.56 & 0.052 & 14.81 & / & OOM \\
\textbf{DPEC} & \textbf{0.017} & \textbf{3.81} & \textbf{0.022} & \textbf{6.40} & \textbf{0.025} & \textbf{13.69} \\
\bottomrule
\end{tabular}%
}
\label{table3}
\end{table}

\newpage 
\begin{figure}[!htp]
    \centering
    \includegraphics[width=0.85\linewidth]{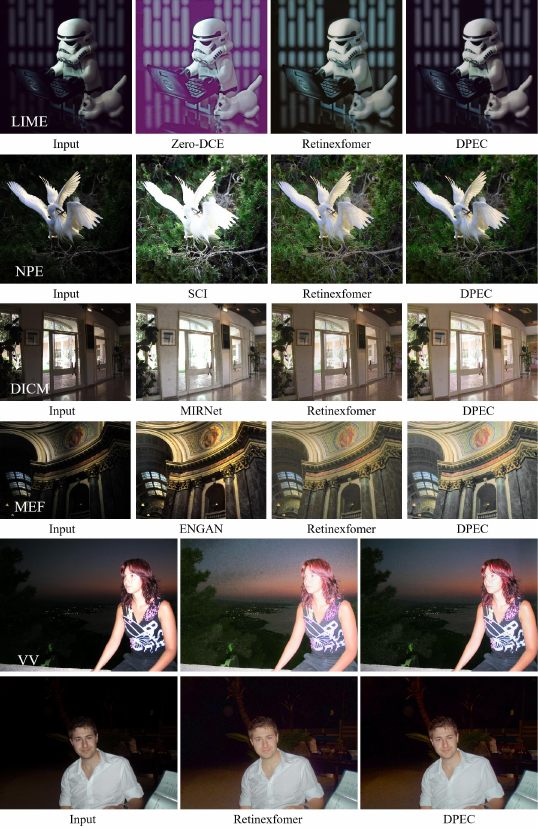}
    \caption{Visual results on the LIME, NPE, MEF, DICM, and VV datasets. Compared to previous algorithms, the results of DPEC exhibit less noise and more natural brightness restoration.}
    \label{fig10}
\end{figure}
\newpage 
\textbf{Qualitative Results:} Visual comparison results of DPEC and other SOTA algorithms are shown in ~\Cref{fig5} and ~\Cref{fig6}, (please enlarge the images for clear details). Compared to previous algorithms, which suffer from color distortion, overexposure/underexposure, excessive noise, overall blurriness, loss of details, and non-natural artifacts, DPEC achieves significant enhancement in overall brightness and better removal of unnatural speckles and artifacts caused by noise. It also exhibits more stable color restoration.

In \Cref{fig7}, we present a visual comparison of the output results from both DPEC-Retinex and DPEC on the LOL dataset. From the enlarged local comparison, it is evident that the DPEC-Retinex outputs, based on the Retinex theory, exhibit greater noise and more pronounced color distortion. Additionally, there is a significant loss of detail in the darker regions of the images.

In ~\Cref{fig10}, we validate the enhancement effectiveness of DPEC in real-world scenarios using datasets without ground truth labels. As a detection scene beyond the labeled dataset, the visual results in ~\Cref{fig10} are more convincing and provide a fairer demonstration of the effectiveness of DPEC. From the visual effects in ~\Cref{fig10}, it can be observed that our method outperforms other SOTA algorithms and unsupervised algorithms in various scenes.

\subsection{Low-light Object Detection}
\label{sec4-3}
\begin{table}[htb]
\centering
\caption{Low-light detection results (map50) on ExDark enhanced by different algorithms. The highest results are in \textcolor{red}{red} color.}
\resizebox{\textwidth}{!}{%
\begin{tabular}{lccccccccccccc}
\toprule
Method & Bicycle & Boat & Bottle & Bus & Car & Cat & Chair & Cup & Dog & Motorbike & People & Table & All $\uparrow$ \\
\midrule
SCI & 65 & \textcolor{red}{67.5} & 54.3 & 77.3 & 70 & 56.8 & \textcolor{red}{56.2} & 52.6 & 46.6 & 47.7 & 70.2 & 50.4 & 59.5 \\
Zero-Dce & 63 & 65.9 & 52.7 & 75.2 & 69.9 & 52.9 & 56.1 & 50.4 & 50.4 & 43.5 & 69.9 & 50 & 58.3 \\
Retinexformer & \textcolor{red}{66.9} & 66.5 & 54.5 & \textcolor{red}{78.4} & 69.8 & 51.5 & 53.6 & 51.8 & 51.6 & \textcolor{red}{49.4} & 70.3 & 52.1 & 59.7 \\
SMNet & 62.2 & 66.9 & 52.1 & 75.4 & \textcolor{red}{70.8} & 51.9 & 55 & \textcolor{red}{56.2} & 50.3 & 49.3 & 70.5 & 49.7 & 59.2 \\
RUAS & 62.8 & 62.2 & 51.4 & 71.5 & 62.2 & 51.5 & 50.1 & 45.6 & 45.1 & 48.4 & 64.5 & 48.3 & 55.3 \\
\textbf{DPEC} & 61.6 & 65.8 & \textcolor{red}{55.5} & 74 & 70.3 & \textcolor{red}{57.7} & 55.3 & 53.2 & \textcolor{red}{52.4} & 48.2 & \textcolor{red}{72} & \textcolor{red}{53.7} & \textcolor{red}{60} \\
\bottomrule
\end{tabular}%
}
\label{table4}
\end{table}

\textbf{Experiment Settings:} We conducted low-light object detection experiments on the Exdark dataset to quantitatively and qualitatively compare the preprocessing effects of different low-light enhancement algorithms on advanced visual tasks. The Exdark dataset consists of a total of 7363 underexposed photos, with 12 object categories annotated with bounding boxes. In the experiments, 5884 images were used for training, and the remaining 1479 images were used for testing. For various low-light enhancement algorithms, we trained them on the LOLv1 dataset and used them as fixed-parameter preprocessing modules to preprocess the Exdark dataset. As for the detector, we used YOLOv8 and fixed the relevant training parameters before training.

\textbf{Quantitative Results:} The map50 scores of different enhancement algorithms are shown in ~\Cref{table4}. Our DPEC achieved the highest map50 score, with 60 AP. It outperformed the best self-supervised algorithm by 0.5 AP and the current best fully supervised algorithm by 0.3 AP. Additionally, DPEC performed the best on five object categories: Bottle, Cat, Dog, People and Table.

\begin{figure}[!h]
    \centering
    \includegraphics[width=0.85\linewidth]{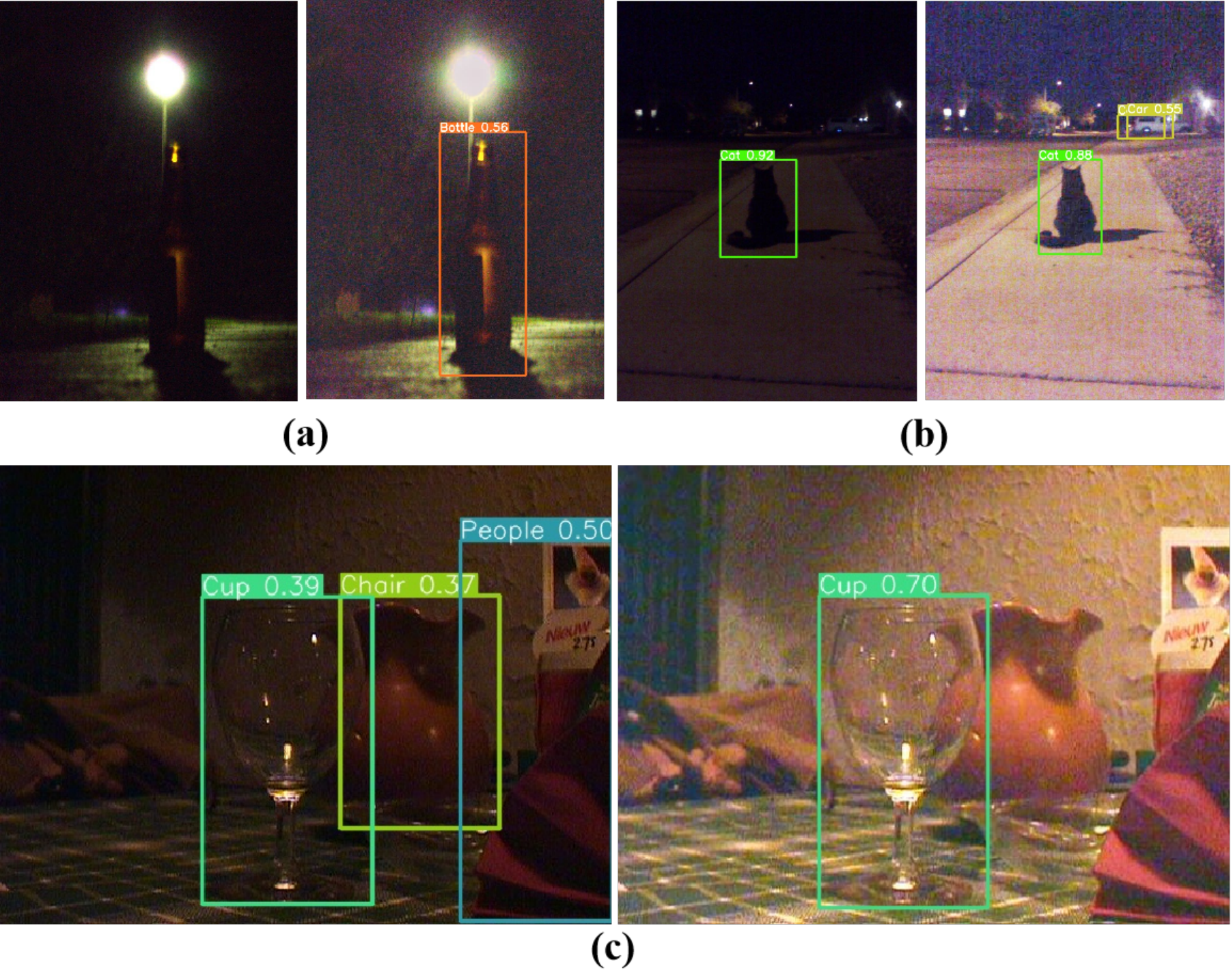}
    \caption{Visual comparison of object detection in low-light (left)
and enhanced (right) scenes by our method on the Exdark dataset. For Figures (a) and (b), the enhanced images reveal more objects concealed in the darkness. In the case of Figure (c), the enhanced image reduces the occurrence of false positives.}
    \label{fig11}
\end{figure}

\textbf{Qualitative Results:} ~\Cref{fig11} shows a visual comparison of detection results in low-light scenes and scenes enhanced by DPEC. In low-light scenes, the detector is prone to missing objects due to underexposure, such as the bottle in ~\Cref{fig11} (a) and the underexposed car in ~\Cref{fig11} (b). And it will also cause misidentification of the object due to insufficient exposure, as shown in ~\Cref{fig11} (c). In contrast, after enhancement by DPEC, the detector can better capture objects in the images and reliably detect them. The anchor boxes are also more accurate, indicating that our method has a certain enhancement effect on high-level visual tasks.
\begin{figure}[!ht]
    \centering
    \includegraphics[width=0.95\linewidth]{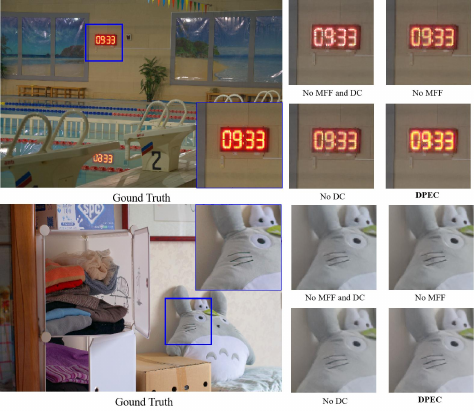}
    \caption{The visual comparison of ablation experiments among different components, it can be seen that after adding the MFF structure and the DC module, there are certain improvement effects on both color and brightness adjustment.}
    \label{fig12}
\end{figure}
\subsection{Ablation Study}
\label{sec4-4}
\textbf{Break-down Ablation:} The ablation experiments performed on the LOL v1 dataset provided a comprehensive evaluation of the performance of various network architectures. By systematically adding or removing specific components, we assessed their individual contributions to the overall image enhancement process. The quantitative results of these experiments are summarized in ~\Cref{table5}. Additionally, ~\Cref{fig12} presents the qualitative visual outcomes corresponding to the ablation studies of the modules integrated within the DPEC structure. These experiments led to the following key insights:

\begin{minipage}{\textwidth}
\begin{minipage}[t]{0.59\textwidth}
\makeatletter\def\@captype{table}
\caption{Break-down ablation to higher performance.}
\resizebox{0.95\textwidth}{!}{%
\begin{tabular}{cccccc}
    \toprule
    Baseline BEE     &MFF           & DC           & PSNR$\uparrow$ &SSIM$\uparrow$ &Parameter\\
    \midrule
    $\checkmark$     &              &              &24.09   &0.844     &1.78M  \\
    $\checkmark$     & $\checkmark$ &              &24.32   &0.846     &2.39M      \\
    $\checkmark$     &              & $\checkmark$ &24.33   &0.853     &1.97M  \\
    \midrule
    $\checkmark$     & $\checkmark$ & $\checkmark$ &\textbf{24.80}   &\textbf{0.855}     &2.58M  \\
    \bottomrule
\end{tabular}%
}
\label{table5}
\end{minipage}
\begin{minipage}[t]{0.35\textwidth}
\makeatletter\def\@captype{table}
\caption{Ablation study of loss function.}
\resizebox{0.95\textwidth}{!}{%
\begin{tabular}{ccc}
    \toprule
    Loss Funcition     & PSNR$\uparrow$     & SSIM$\uparrow$ \\
    \midrule
    W/o $\mathcal{L}_{ssim}$  & 25.845  & 0.936    \\
    W/o $\mathcal{L}_{per}$     & 23.520        & 0.931      \\
    W/o $\mathcal{L}_{his}$     & 26.272       & 0.943  \\
    W/o $\mathcal{L}_{inner}$     & 26.278      & 0.944  \\
    W/o $\mathcal{L}_{tv}$     & 26.078       & 0.939  \\
    W/o $\mathcal{L}_{smooth}$     & 26.094       & 0.941  \\
    \midrule
    \textbf{DPEC} &\textcolor{red}{26.591} & \textcolor{red}{0.942}\\
    \bottomrule
\end{tabular}%
}
\label{table6}
\end{minipage}
\end{minipage}
\\
\begin{itemize}
    \item The original BBE baseline model demonstrates exceptional performance in basic image enhancement, achieving PSNR and SSIM metrics of 24.09 dB and 0.844, respectively.
    \item The integrated Multi-scale Feature Fusion (MFF) structure significantly optimizes the processing capability of the two-layer BBE, enhancing the model's ability to capture fine details. Compared to the original BBE, there is an improvement of 0.23 dB in PSNR.
    \item The final model, DPEC, was obtained by combining DenoiseCNN with the BBE model integrated with the MFF structure.  We found that the introduction of the denoising module positively impacts overall image enhancement, further improving image clarity and visual quality.  DPEC achieves the highest scores in both PSNR and SSIM, reaching 24.80 dB and 0.855, respectively.
\end{itemize}

Considering all the results from the ablation experiments, our DPEC model achieved the highest performance in terms of both PSNR and SSIM metrics. Compared to the original BBE baseline model, the BBE with the integration of the multi-scale feature fusion structure was able to capture edge and detail information more finely. With the introduction of DenoiseCNN, the noise issue that was not yet removed in the original input was successfully addressed, resulting in a significant improvement in image quality.

\textbf{Configuration of Losses:} As shown in \Cref{table6}, we validated the effectiveness of various loss functions using the LOLv2-syn dataset. The ablation experiments demonstrated that removing the SSIM loss decreased both PSNR and SSIM, confirming its role in maintaining image quality, while the absence of the Perceptual Loss significantly reduced PSNR due to the loss of contextual semantic information. The Histogram Loss was crucial for adjusting brightness and color texture, improving PSNR with minimal impact on SSIM. Additionally, the TV loss contributed to noise reduction and edge preservation, the inner loss helped maintain structural fidelity, and the smooth loss ensured smooth transitions and reduced artifacts, as their removal led to slight declines in PSNR. Together, these losses complement the core loss functions, enabling DPEC to achieve robust and high-quality image enhancement.

\section{Conclusion and Limitations}
\subsection{Limitations of the DPEC Method}
\label{sec5-1}
Based on our findings, the DPEC method exhibits several limitations in illumination adjustment. While it performs well in color preservation and noise control, its reliance on the multiplicative decomposition of Retinex theory (i.e., \( S = I \odot R \)) limits its adaptability to scenarios lacking well-lit reference images. Specifically, the multiplicative constraints between the illumination map \( I \) and reflectance \( R \) lead to non-convex optimization challenges, particularly when paired with unpaired training data. To address this, future improvements could employ logarithmic transformations to convert multiplicative operations into additive ones (i.e., \( \log(S) = \log(I) + \log(R) \)). This transformation simplifies gradient-based optimization while retaining the physical interpretability of the Retinex model. Additionally, DPEC may introduce artifacts or over-smoothing in regions with extremely low illumination due to the limitations of error estimation in capturing abrupt lighting variations. In future work, a frequency-sensitive refinement module could be integrated to preserve critical edge information during illumination adjustment and further enhance its performance.

\subsection{Summary and Future Directions}
\label{sec5-2}

This study presents DPEC, a novel low-light image enhancement framework leveraging the Vision Mamba architecture. By addressing the limitations of traditional Retinex-based methods, DPEC introduces a dual-path error compensation mechanism and a dedicated denoising module. The Vision Mamba architecture facilitates efficient long-range dependency modeling, while the HIS-Retinex Loss ensures natural brightness distribution and accurate color recovery. Experiments on the LOL and LSRW datasets demonstrate that DPEC achieves superior image quality with significantly reduced computational costs, offering a 2.5$\times$ faster inference speed than RetinexFormer at 1080P resolution, making it highly suitable for real-time applications.

Future work will focus on enhancing DPEC's generalization to diverse lighting conditions without reliance on reference images and exploring advanced error estimation techniques for extreme low-light and complex lighting scenarios. Additionally, optimizing DPEC for deployment on edge devices, such as FPGAs, will be prioritized to expand its real-world applicability.

To support further research, the code and datasets are publicly available.

\textbf{Declarations} Conflict of interest: The authors declare they have no conflict of interest.





\end{document}